\documentclass{article}

\usepackage[final]{neurips_2022}

\usepackage[utf8]{inputenc} % allow utf-8 input
\usepackage[T1]{fontenc}    % use 8-bit T1 fonts
\usepackage{hyperref}       % hyperlinks
\usepackage{url}            % simple URL typesetting
\usepackage{booktabs}       % professional-quality tables
\usepackage{amsfonts}       % blackboard math symbols
\usepackage{amsmath}
\usepackage{nicefrac}       % compact symbols for 1/2, etc.
\usepackage{microtype}      % microtypography
\usepackage{xcolor}         % colors
\usepackage{graphicx}
\usepackage{subfig}

\usepackage{tabularx}

\title{Neural Architecture for Online Ensemble Continual Learning}

\author{%
Mateusz Wójcik$^{1, 2}$ \quad Witold Kościukiewicz$^{1, 2}$ \quad Tomasz Kajdanowicz$^1$ \quad Adam Gonczarek$^2$ \\
$^1$Wroclaw University of Science and Technology \quad $^2$Alphamoon Ltd., Wrocław\\
\texttt{\{mateusz.wojcik,witold.kosciukiewicz,tomasz.kajdanowicz\}@pwr.edu.pl}\\
\texttt{adam.gonczarek@alphamoon.ai}\\
}

\begin{document}

\maketitle

\begin{abstract}
Continual learning with an increasing number of classes is a challenging task. The difficulty rises when each example is presented exactly once, which requires the model to learn online. Recent methods with classic parameter optimization procedures have been shown to struggle in such setups or have limitations like non-differentiable components or memory buffers. For this reason, we present the fully differentiable ensemble method that allows us to efficiently train an ensemble of neural networks in the end-to-end regime. The proposed technique achieves SOTA results without a memory buffer and clearly outperforms the reference methods. The conducted experiments have also shown a significant increase in the performance for small ensembles, which demonstrates the capability of obtaining relatively high classification accuracy with a reduced number of classifiers.
\end{abstract}

\section{Introduction}

Over the last few years, neural networks have become a widely used and effective tool, especially in supervised learning problems \cite{resnet,devlin2018bert,rawat2017deep}. The parameter optimization process based on a gradient descent works well when the data set is sufficiently large and available entirely during the training process. Otherwise, the catastrophic forgetting \cite{french1999catastrophicforgetting} will occur, which makes neural networks unable to be trained incrementally. The field of continual learning aims to develop methods that enable the accumulation of new knowledge without forgetting previously inferred one. 

Currently, the methods that are guaranteed to be most effective across various tasks utilize a memory buffer \cite{CVPR-2020-competition}. While this is a relatively simple and effective approach, it requires constant access to the data. In many practical real-world applications, this disqualifies such methods due to privacy policies or data size \cite{Safety1}. It has also been shown that methods without a memory buffer are not effective in class incremental \cite{3scenarios} setup with classic optimization algorithms like e.g. Adam \cite{mirzadeh2020understanding}.

In this paper, we present a fully differentiable neural architecture for online class incremental continual learning called DE\&E. The architecture is inspired by an Encoders and Ensembles (hereafter referred to as E\&E) \cite{reference_method} and adapted to the most challenging task-free online class incremental setup. 
%We introduce a \textit{soft KNN} layer, we proposed a novel ensemble predictions weighting strategy and we show effectiveness of the presented architecture. 
Our method retains advantages of E\&E while increasing its accuracy, reducing forgetting, enables end-to-end ensemble training and significantly improving performance when number of parameters is low (small ensembles). We demonstrate that the proposed architecture achieves SOTA results in evaluated scenarios. In summary, our contributions are as follows: 1) we introduced a differentiable KNN layer \cite{Diff-KNN-SOFT} into the model architecture, 2) we proposed a novel approach to aggregate classifier predictions in the ensemble, 3) we demonstrate the proposed architecture effectiveness by achieving SOTA results on popular continual learning benchmarks without a memory buffer.

\section{Model architecture}

\begin{figure}[htp]
    \centering
    \includegraphics[width=\textwidth]{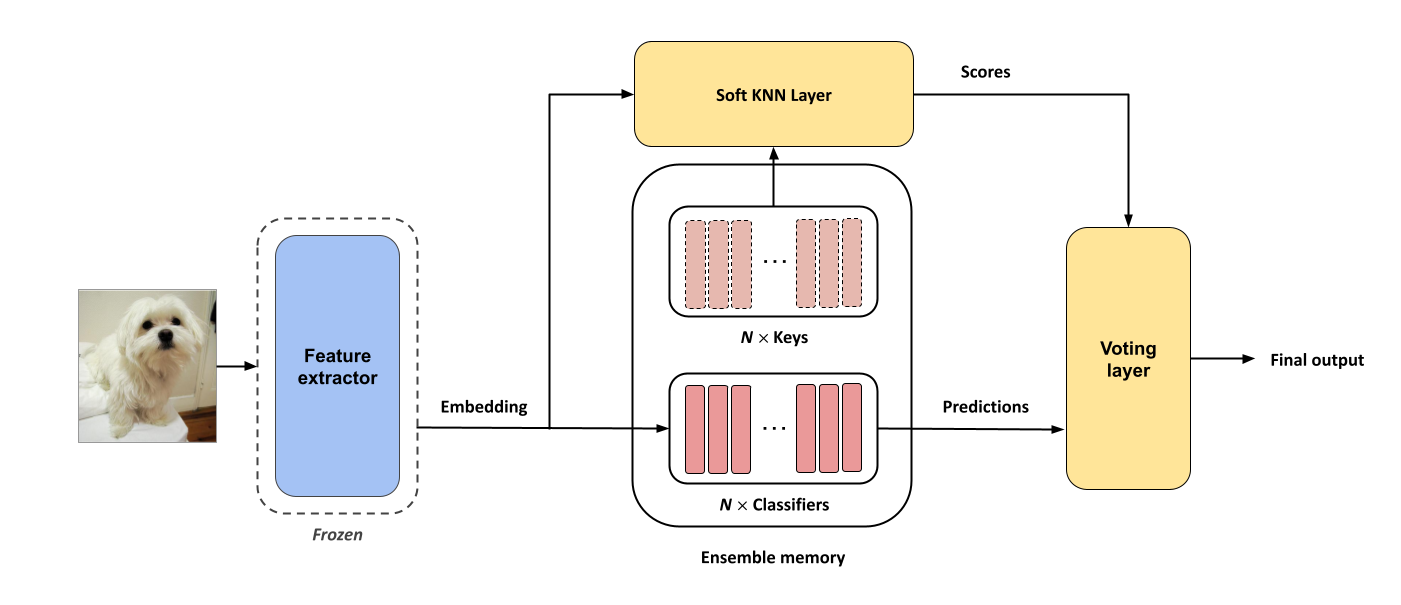}
    \caption{Architecture of the proposed model. Input image is processed by the feature extractor. Obtained embeddings are used to find the most relevant classifiers according to assigned keys. The \textit{soft KNN} layer approximates the \textit{soft KNN} scores. Predictions are weighted in the voting layer by both cosine similarity and \textit{soft KNN} scores. Final output is the class with the highest voting score.}
    \label{fig:architecture_e2e}
\end{figure}

%The model consists of a pre-trained feature extractor, an ensemble of single-layer neural networks with keys assigned, and two fully differentiable modules (\textit{soft KNN} layer and voting layer). The input data is passed into a feature extractor that returns feature embeddings. Then, each classifier predicts the output based on the extracted features. Relative to the reference method, the module responsible for selecting the best matching classifiers for the input example has been replaced by a differentiable layer (\textit{soft KNN}).
%This enables us to train parts of the model that were frozen until now ([cite appendix]).
%Additionally, we adopted the approximations from \textit{soft KNN} layer as additional weights during ensemble voting. The full architecture is presented in Figure \ref{fig:architecture_e2e}. 

\paragraph{Feature extractor.}
The full model architecture is presented in Figure \ref{fig:architecture_e2e}. The first component of the proposed architecture is a multi-layer feature extractor that transforms input data into the embedding space. It can be described by the following mapping $\mathbf{z}=F(\mathbf{x})$, where $\mathbf{x}\in\mathbb{R}^{D}$ is an input example and $\mathbf{z}\in\mathbb{R}^{M}$ is a $M$-dimensional embedding. The approach we follow assumes the use of a pre-trained model with frozen parameters. Such a procedure makes it possible to completely prevent the extractor from forgetting by isolating feature space learning from the classification process. 

%This guarantees stability in obtaining features \cite{reference_method} and isolate . reducing the risk of observing a recency bias \cite{RecencyBias}.

\paragraph{Keys and classifiers.}
We use an ensemble of $N$ classifiers $f_{n}(\cdot)$, where each of them maps the embedding into a $K$-dimensional output vector $\mathbf{\hat{y}}_{n}=f_{n}(\mathbf{z})$. With each classifier, there is an associated key vector $\mathbf{k}_{n}\in\mathbb{R}^{M}$ with the same dimensionality as the embedding. The keys help to select the most suitable models for specialization with respect to the currently processed input example. They are initialized randomly from normal distribution. We use simple single-layer neural networks as classifiers, with fan-in variance scaling as the weight initialization strategy. The network output is activated by a hyperbolic tangent function (\textit{tanh}).

\paragraph{Soft $\kappa$-nearest neighbors layer.}
The standard KNN algorithm is often implemented using ordinary sorting operations that make it impossible to determine the partial derivatives with respect to the input. It removes the ability to use KNN as part of end-to-end neural models. However, it is possible to obtain a differentiable approximation of the KNN model by solving the Optimal Transport Problem \cite{optimaltransportproblem}. Based on this concept, we add a differentiable layer to the model architecture. We call this layer soft $\kappa$-nearest neighbors (\textit{soft KNN}).
In order to determine the KNN approximation, we first compute a cosine distance vector $\mathbf{c}\in \mathbb{R}^{N}$ between the embedding and the keys:
\begin{equation}
    c_{n} = 1-\cos(\mathbf{z},\mathbf{k}_{n}),
\end{equation}
where $\mathbf{\cos(\cdot,\cdot)}$ denotes the cosine similarity. Next, we follow the idea of a soft top-$\kappa$ operator presented in \cite{Diff-KNN-SOFT}, where $\kappa$ denotes the number of nearest neighbors. Let $\mathbf{E}\in\mathbb{R}^{N\times 2}$ be the Euclidean distance matrix with the following elements:
\begin{equation}
e_{n,0}=(c_{n})^{2},\ \ \ e_{n,1}=(c_{n}-1)^{2}.    
\end{equation}
And let $\mathbf{G}\in\mathbb{R}^{N\times 2}$ denote the similarity matrix obtained by applying the Gaussian kernel to $\mathbf{E}$:
\begin{equation}
\mathbf{G}= \exp(-\mathbf{E}/\sigma),    
\end{equation}
where $\sigma$ denotes the kernel width. The $\exp$ operators are applied elementwise to matrix $\mathbf{E}$. 

\newpage

We then use the Bregman method, an algorithm designed to solve convex constraint optimization problems, to compute $L$ iterations of Bregman projections in order to approximate their stationary points:
\begin{equation}
    \mathbf{p}^{(l+1)}=\frac{\boldsymbol\mu}{\mathbf{G}\mathbf{q}^{(l)}},\ \ \ \mathbf{q}^{(l+1)}=\frac{\boldsymbol\nu}{\mathbf{G}^{\top}\mathbf{p}^{(l+1)}},\ \ \ l=0,\dots,L-1
\end{equation}
where $\boldsymbol\mu=\mathbf{1}_{N}/N$, $\boldsymbol\nu=[\kappa/N,(N-\kappa)/N]^{\top}$, $\mathbf{q}^{(0)}=\mathbf{1}_{2}/2$, and $\mathbf{1}_{i}$ denotes the $i$-element all-ones vector. 
Finally, let $\boldsymbol\Gamma$ denotes the optimal transport plan matrix and is given by:
\begin{equation}
% \boldsymbol\G
\boldsymbol\Gamma = \mathrm{diag}(\mathbf{p}^{(L)})\cdot \mathbf{G} \cdot \mathrm{diag}(\mathbf{q}^{(L)})  
\end{equation}
As the final result $\boldsymbol\gamma\in {\mathbb{R}^{N}}$ of the soft $\kappa$-nearest neighbor operator, we take the second column of $\boldsymbol\Gamma$ multiplied by $N$ i.e. $\boldsymbol\gamma=N  \boldsymbol\Gamma_{:,2}$. $\boldsymbol\gamma$ is a soft approximation of a zero-one vector that indicates which $\kappa$ out of $N$ instances are the nearest neighbors. Introducing the \textit{soft KNN} enables us to train parts of the model that were frozen until now (\ref{appendix:section:trainable-keys}).

\paragraph{Voting layer.}
We use both $c_{n}$ and $\boldsymbol\gamma$ to weight the predictions by giving the higher impact for classifiers with keys similar to extracted features. The obtained approximation $\boldsymbol\gamma$ has two main functionalities. It eliminates the predictions from classifiers outside $\kappa$ nearest and weights the result. Since the Bregman method does not always completely converge, the vector $\kappa$ contains continuous values that are close to 1 for the most relevant classifiers. We make use of this property during the ensemble voting procedure. The higher the $\kappa$ value for a single classifier, the higher its contribution toward the final ensemble decision. The final prediction is obtained as follows:

\begin{equation}
\mathbf{\hat{y}}=\frac{\sum_{n=1}^{N} \gamma_{n}c_{n}\mathbf{\hat{y}}_{n}}{\sum_{n=1}^{N}c_{n}}    
\end{equation}

\paragraph{Training}
To effectively optimize the model parameters, we follow the training procedure presented in \cite{reference_method}. It assumes the use of a specific loss function that is the inner product between the ensemble prediction and the one-hot coded label: 

\begin{equation}
\mathcal{L}(\mathbf{y}, \hat{\mathbf{y}})=-\mathbf{y}^{\top} \hat{\mathbf{y}}
\end{equation}

Optimizing this criterion yields an advantage of using a \textit{tanh} activation function, significantly reducing catastrophic forgetting. Following the reference method, we also use an optimizer that discards the value of the gradient and uses only its sign to determine the update direction. As a result, the parameters are being changed by a fixed step during the training.

\section{Experiments}

\paragraph{Setup}

\paragraph{Results.} The results of the evaluation on MNIST and CIFAR-10 are presented in Table \ref{tab:small-ensemble}. For all setups evaluated, our model performed best improving results of main reference method (E\&E) up to 6\%. 
We can also see a significant difference in achieved accuracy between the DE\&E approach and baselines.
%The proposed architecture proved to be superior, although the improvement over the reference method is relatively small. 
Furthermore, it achieved this results without replaying training examples seen in the past, making it more practical relative to memory based methods (Replay, A-GEM, GEM) with 10 examples stored per experience (one split). For the ensemble of 128 classifiers and MNIST, our architecture achieved results more than 18\% better than the best method with a memory buffer.
%Experiments show how the task-free protocol with an increasing number of classes is a difficult problem. As the literature indicates, many continual learning methods perform well in easier approaches where the tasks are clearly defined and additionally the subset of classes at prediction is narrowed by providing a task-id (multi-head evaluation) \cite{3scenarios}.
%Importantly, our study indicates that the introduction of a differentiable KNN layer along with the proposed weighting mechanism achieves results in all cases better than the baseline classifier ensemble. 
%Thus, we gain a fully differentiable neural architecture with comparable performance for large ensembles.

\begin{figure}
\centering
\begin{minipage}{.46\textwidth}  \centering
  \includegraphics[width=\linewidth]{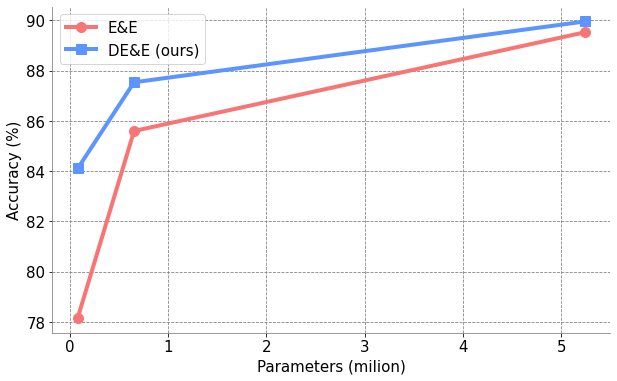}
  \captionof{figure}{Number of weights in ensembles (16, 128, 1024 classifiers) and achieved accuracy (\%) on 10-split MNIST.}
  \label{fig:parameters}
\end{minipage}%
\hspace{1cm}
\begin{minipage}{.46\textwidth}
  \centering
  \includegraphics[width=\linewidth]{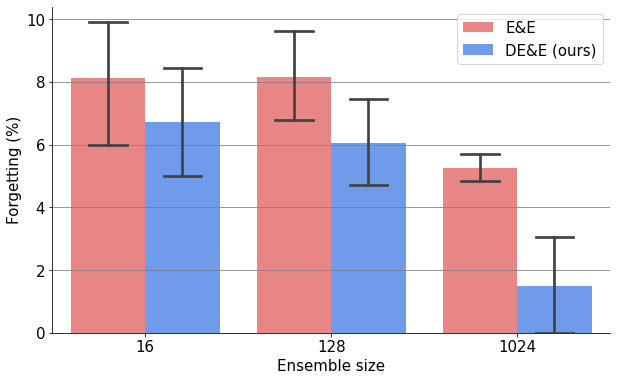}
  \captionof{figure}{Averaged forgetting rate (the lower the better) for ensembles evaluated on 10-split MNIST.}
  \label{fig:forgetting}
\end{minipage}
\end{figure}

\begin{table}[!ht]
    \caption{Accuracy (\%) and standard deviation for methods evaluated on MNIST and CIFAR-10. For experimental setup details see \ref{appendix:section:details}.}
    \label{tab:small-ensemble}
    % \fontsize{6}{6}\selectfont
    % \renewcommand{\arraystretch}{1.5}
    \centering
        \begin{tabular*}{\textwidth}{@{\extracolsep{\fill}} lcccc @{}}
    \toprule
                        & \multicolumn{2}{c}{MNIST (10 splits)} & \multicolumn{2}{c}{CIFAR-10 (5 splits)} \\ \cmidrule(l){2-5} 
                        & $N = 16$          & $N = 128$         & $N = 64$            & $N = 128$           \\ \midrule

    Naive            & $14.41 \pm 5.99$  & $11.63 \pm 2.22$  & $19.65 \pm 0.33$   & $19.70 \pm 0.36$    \\
    LwF \cite{LWF}   & $12.38 \pm 3.99$  & $9.88 \pm 0.55$  & $19.48 \pm 0.55$   & $19.62 \pm 0.60$    \\
    EWC \cite{EWC}   & $14.33 \pm 4.44$  & $10.97 \pm 2.32$  & $19.52 \pm 0.29$   & $19.88 \pm 0.50$    \\
    SI \cite{zenke2017continual}              & $10.18 \pm 1.00$  & $17.22 \pm 4.64$  & $17.97 \pm 2.40$   & $21.32 \pm 5.76$    \\
    CWR* \cite{lomonaco2017core50}            & $16.41 \pm 5.42$  & $10.38 \pm 0.79$  & $18.92 \pm 2.97$   & $22.41 \pm 2.00$    \\
    \midrule
    GEM \cite{GEM} (10 / exp)   & $67.81 \pm 2.61$  & $58.92 \pm 6.34$  & $30.75 \pm 1.47$   & $29.27 \pm 1.46$    \\
    A-GEM \cite{AGEM} (10 / exp)           & $53.59 \pm 5.21$  & $21.31 \pm 15.90$  & $39.86 \pm 14.25$   & $36.12 \pm 6.19$    \\
    Replay \cite{TinyEpisodicMemory} (10 / exp)           & $74.49 \pm 3.84$  & $69.02 \pm 4.90$  & $44.03 \pm 3.72$   & $43.82 \pm 7.10$    \\
    \midrule
    % Vanilla          & ?  & ?  & ?   & ?    \\
    % Tanh             & ?  & ?  & ?   & ?    \\
    E\&E \cite{reference_method}            & $78.16 \pm 1.85$  & $85.60 \pm 0.52$  & $46.34 \pm 1.98$    & $56.24 \pm 1.41$     \\
    DE\&E (ours)     & $\mathbf{84.19 \pm 1.00}$  & $\mathbf{87.54 \pm 0.24}$  & $\mathbf{48.78 \pm 1.34}$    & $\mathbf{59.36 \pm 0.73}$    \\ \bottomrule
    \end{tabular*}
\end{table}

In addition, we observed that the proposed method significantly improves the performance of small ensembles. The smaller the ensembles, the higher the gain in accuracy. For MNIST and the ensemble of 16 models, the improvement was up to approximately 6\% over the E\&E. For the 64 classifiers and CIFAR-10, the improvement was about 5\%. Figure \ref{fig:parameters} shows the comparison of the total number of weights of ensembles of different sizes and the achieved classification performance. The proposed method achieves higher results having the same number of parameters. 
%In the case of evaluation on the MNIST data set, the number of parameters is less, the gain of using the end-to-end architecture is more significant. 
For an ensemble of 1024 classifiers, the accuracy is already very close, suggesting that the gain decreases with large ensembles.

\newpage

An important advantage of the proposed method is a low forgetting rate \cite{chaudhry2018riemannian}.
% TODO move it to appendix 
% We define forgetting as the difference between the expected accuracy of the model on a certain task (just after training) and the accuracy after finishing learning on all other tasks \cite{delange2021continual}. The expected accuracy of the model is equal to that immediately after the end of training on that task.
We observed significantly reduced forgetting relative to the reference method, as shown in Figure \ref{fig:forgetting}. Stronger specialization amplified by the introduced voting method makes classifiers less likely to lose acquired knowledge. The larger the ensemble the relatively less knowledge is forgotten. Empirically, this phenomenon can be explained by the fact that a larger ensemble means better coverage of the key data space, making the models specialize in classifying specific groups of examples. As a result, we protect models against domain shift of input examples, thus making them easier to classify and harder to forget.
% The forgetting phenomenon is also evident when tracking accuracy on individual tasks during consecutive training in class incremental setup. 
% The smaller ensemble acquires less knowledge on subsequent tasks and forgets much more.

\section{Conclusions}

In this paper, we proposed a neural architecture for online continual learning with training procedure specialized in challenging class incremental problems. 
%Our method extends the Encoders and Ensembles approach and is adapted for challenging class incremental problems.
The presented architecture introduces a fully differentiable \textit{soft KNN} layer and novel prediction weighting strategy based on the \textit{soft KNN}. This components amplified the influence of most specialized classifiers on the final prediction. 
% used to select appropriate classifiers for specialization
%We performed evaluations on the MNIST, CIFAR-10 and CIFAR-100 data sets in various classification scenarios with an increasing number of classes.
As a result, we showed improved accuracy for all of the cases studied and achieved SOTA results. We have shown that it is possible to noticeably improve the quality of classification
%(+5\% for an ensemble of 16 on 10-split MNIST) 
using the proposed techniques and this effect is observed especially in small ensembles that gained significantly higher performance.
%Training neural networks using the classical gradient descent algorithm is a time-consuming process for large models. The introduced \textit{soft KNN} layer forces to perform more complex computations that reflects on both forward and backward passes time. %Furthermore, note that the training time for such models depends linearly on the number of models in the ensemble. 
%Popular image classification tasks are not very demanding in this regard, but for various domains it can be a major drawback. Especially since the quality of the encoder directly affects the accuracy of the classification and can reduce the chances for satisfactory results despite a well-designed model architecture. However, robust multimodal models \cite{Modality1,Modality2} are now rapidly developing, which brings hope for solving this problem. It is also worth noting the potential of using an end-to-end architecture that allows possible unfreezing and fine-tuning the connected feature extractor. However, with the current state of knowledge, it is necessary to have a frozen encoder that, due to the still unsolved problem of catastrophic forgetting in deep learning, plays a key role in the training process. Undoubtedly, the field of continual learning using ensemble methods needs more attention due to its vast potential.
As a result, the presented architecture outperforms methods with memory buffer and enables researchers to make further steps towards overrun the current SOTA in class incremental problems.

\section*{Acknowledgement}
The research was conducted under the Implementation Doctorate programme of Polish Ministry of Science and Higher Education and also partially funded by Department of Artificial Intelligence. It was also partially co-funded by the European Regional Development Fund within the Priority Axis 1 “Enterprises and innovation”, Measure 1.2. “Innovative enterprises, sub-measure 1.2.1. “Innovative enterprises – horizontal competition” as part of ROP WD 2014-2020, support contract no. RPDS.01.02.01-02-0063/20-00.

% \newpage

% \section*{References}

\bibliographystyle{abbrv}
\bibliography{bibliography.bib}

% End references

\appendix

\section{Appendix}

\subsection{Code}

Code is currently available in \href{https://github.com/mateusz-wojcik-97/neural-architecture-for-online-ensemble-cl}{Github repository} (\url{https://github.com/mateusz-wojcik-97/neural-architecture-for-online-ensemble-cl}).

\subsection{Implementation details.}\label{appendix:section:details}

We use PyTorch to both reproduce the E\&E results and implement the DE\&E method. We use a pre-trained ResNet-50 model as the feature extractor for the CIFAR-10 data set. The model is available in the following GitHub repository, \url{https://github.com/yaox12/BYOL-PyTorch}, and is used under MIT Licence. For MNIST, we trained a variational autoencoder on the Omniglot data set. We based our implementation of the \textit{soft KNN} layer on the code provided with \url{https://proceedings.neurips.cc/paper/2020/hash/ec24a54d62ce57ba93a531b460fa8d18-Abstract.html}. All data sets used are public.

\paragraph{Data sets.}

For model evaluation, we used two popular data sets: MNIST, CIFAR-10 and we also perform some additional experiments using the CIFAR-100 dataset (Sections \ref{appendix:section:complexity}, \ref{appendix:section:large-and-100}). The selected data sets are characterized by varying difficulty. MNIST provides images that are significantly easier to classify due to their simple structure. In contrast, CIFAR-10 data set contain slightly larger, color images and provides 10 classes. Each data set was tested using two commonly used configurations: The first one covered a class incremental scenario where each class appears separately (one at a time). The second scenario involved multiple classes appearing at once. Depending on the data set, these configurations varied - for MNIST and CIFAR-10, the 10-split and 5-split approaches were evaluated. By split we mean into how many parts the data set labels were divided. % Thus, 10-split for MNIST and CIFAR-10 are fully class incremental setup.
It is worth noting that the 5-split approach can be treated as task-incremental by default when task-id is provided \cite{3scenarios}. However, during our research it was taught without providing a task identifier to the model (but we conventionally refer to them as tasks). Such a procedure makes the task even more complicated, because the choice must always be made between all classes. 

\paragraph{Feature extractor.}

\begin{figure}[!th]
    \centering
    \includegraphics[width=\textwidth]{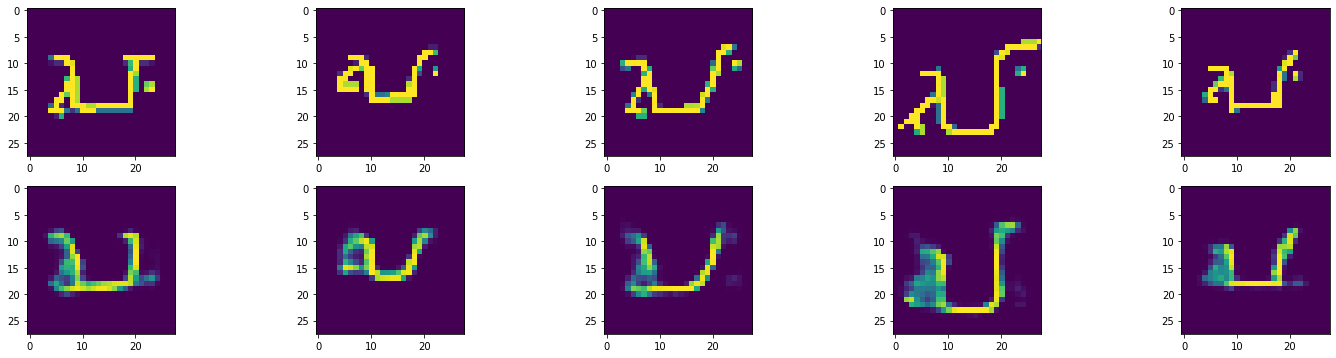}
    \caption{Examples of image reconstruction using a trained VAE. The top row contains the original images from Omniglot while the bottom row shows the reconstructions.}
    \label{fig:vae_example}
\end{figure}

To obtain the features for MNIST, a variational autoencoder (VAE) was trained on the Omniglot data set \cite{omniglot}. The VAE architecture follows the one introduced with the E\&E method. The VAE feature vector has a size of 512. The training setup is presented in Table \ref{tab:vae}. To train the VAE, we used the standard reconstruction loss,

\begin{equation}
\|\mathbf{x} - \mathbf{\hat{x}}\|^2 + \beta KL(\mathbf{z}, \mathcal{N}(0, \mathbf{I}))
\end{equation}

where $\mathbf{\hat{x}}$ is the decoder output, $\mathcal{N}(0, \mathbf{I})$ is the standard normal distribution, and $\beta$ is the regularization term for latent loss. We trained the model until the VAE reconstruction loss exceeded the fixed threshold. After training was complete, we froze the encoder weights and extracted them from the trained VAE. Examples of image reconstruction using the trained model are shown in Figure \ref{fig:vae_example}.

\begin{table}[!ht]
    \caption{Hyperparameters of the VAE trained on the Omniglot dataset.}
    \label{tab:vae}
    \centering
    \begin{tabular*}{0.3\textwidth}{ll}
    \toprule
                        
    Hyperparameter          & Value    \\ \midrule
    Learning rate  & 0.001              \\
    Encoding size  & 512              \\
    Input size  & 28              \\
    Batch size  & 48             \\ 
    Loss threshold & 0.02                  \\
    $\beta$ & 0.001                  \\ \bottomrule
    \end{tabular*}
\end{table}

In order to extract high quality features from CIFAR-10, it was necessary to train a more complex model beforehand. The Resnet-50 model \cite{resnet} trained on ImageNet data \cite{ImageNet} was used. It produces a feature vector with a size 2048. The four-times larger feature vector is necessary to provide quality features for classifying images from such data sets. The BYOL technique \cite{grill2020bootstrap} was utilized in order to obtain more informative embeddings. BYOL is an approach dedicated to learning image representations using a self-supervised approach that involves the use of two neural networks. The first, called an online network, learns to predict the representation generated by the second network, called a target. Both networks receive the same image as an input, but with two different transformations applied. The contrastive loss is minimized and the target network parameters are updated as the slow-moving average of the online network.

\paragraph{Baselines.}
We use Naive, LwF \cite{LWF}, EWC \cite{EWC}, SI \cite{zenke2017continual}, CWR* \cite{lomonaco2017core50}, GEM \cite{GEM}, A-GEM \cite{AGEM} and Replay \cite{TinyEpisodicMemory} approaches as baselines to compare with our method. We utilize the implementation from Avalanche (\url{https://avalanche.continualai.org/}), a library designed for continual learning tasks. The main purpose of this comparison was to determine how the proposed method performs against classical approaches and, in particular, against the methods with memory buffer, which gives a significant advantage in class incremental problems. The recommended hyperparameters for each baseline method vary across usages in literature, so we chose them based on our own internal experiments. For a clarity, we keep hyperparameter naming nomenclature from the Avalnache library. For EWC we use $lambda$ = $10000$. The LwF model was trained with $alpha$ = $0.15$ and $temperature$ = $1.5$. For SI strategy we use $lambda$ = $5e7$ and $eps$ = $1e-7$. The hyperparameters of the memory based approach GEM were set as follows: $memory\_strength$ = $0.5$, $patterns\_per\_exp$ = $10$, which implies that with every experience (split), 10 examples will we accumulated. This has a particular importance when the number of classes is large. With this setup and 10-split MNIST, memory contains 100 examples after training on all classes. Having a large memory buffer makes achieving high accuracy much easier. For the A-GEM method use the same number of examples in memory and $sample\_size$ = $10$. All models were trained using Adam optimizer with a $learning\_rate$ of $0.0005$ and $batch\_size$ of $32$. We chose cross entropy as a loss function and performed one training epoch for each experience. To fairly compare baseline methods with ensembles, as a backbone we use neural network with a similar number of parameters (as in ensemble). Network architectures for each experimental setup are shown in Table \ref{tab:parameters-comparison}. All baseline models were trained by providing embeddings produced by feature extractor as an input.

\begin{table}[!ht]
    \caption{Architecture of neural networks used as backbones for baseline models depends on experimental setup.}
    \label{tab:parameters-comparison}
    \centering
    \begin{tabular*}{0.47\textwidth}{lll}
    \toprule
                        
    Dataset & Classifiers         & Network layers    \\ \midrule
    MNIST & 16   & [512, 157, 10]              \\
    MNIST & 128 & [512, 1256, 10]                  \\
    CIFAR-10 & 64 & [2048, 637, 10]                   \\
    CIFAR-10 & 128 & [2048, 1274, 10]                   \\ \bottomrule
    \end{tabular*}
\end{table}

We used E\&E \cite{reference_method} as the main reference method. It uses an architecture similar to that of a classifier ensemble, however the nearest neighbor selection mechanism itself is not a differentiable component and the weighting strategy is different. In order to reliably compare the performance, the experimental results of the reference method were fully reproduced. Both the reference method and the proposed method used exactly the same feature extractors. Thus, we ensured that the final performance is not affected by the varying quality of the extractor, but only depends on the solutions used in the model architecture and learning method.

\paragraph{Ensembles.}

Both E\&E and our DE\&E were trained with the same set of hyperparameters, excluding hyperparameters in the \textit{soft KNN} layer. The setup is shown in Table \ref{tab:soft-knn}. Every experiment was performed in an online manner, which means one example is shown to the model only once. We use ensembles of sizes 16, 64, 128 and 1024. Based on the size, ensembles have various number of nearest neighbors assigned (Table \ref{tab:knn}). Depends on the data set, the input batch size was different. For MNIST the batch size was 60. In contrast, for CIFAR-10 we use the batch size of 10 due to larger embedding vector produced by the feature extractor.

The keys for classifiers in ensembles were randomly chosen from the standard normal distribution and normalized using the $L2$ norm. The same normalization was applied to encoded inputs during lookup for matching keys. We used Adam optimizer with a learning rate of 0.0005 to train the keys.

\begin{table}[!ht]
    \caption{Hyperparameters of the DE\&E model.}
    \label{tab:soft-knn}
    \centering
    \begin{tabular*}{0.3\textwidth}{ll}
    \toprule
                        
    Hyperparameter          & Value    \\ \midrule
    Learning rate  & 0.0001              \\
    Weight decay  & 0.0001              \\
    \textit{Tanh} scaling  & 250              \\ \bottomrule
    $\sigma$  & 0.0005              \\ 
    $L$ & 400                  \\ \bottomrule
    \end{tabular*}
\end{table}

\begin{table}[!ht]
    \caption{Number of neighbors used for each evaluated ensemble size.}
    \label{tab:knn}
    \centering
    \begin{tabular*}{0.37\textwidth}{ll}
    \toprule
                        
    $N$ (ensemble size)          & $\kappa$ (neighbors)    \\ \midrule
    16  & 4              \\
    64 & 8                  \\
    128 & 16                  \\
    1024 & 32                  \\ \bottomrule
    \end{tabular*}
\end{table}

\paragraph{Soft KNN.} 

We use the Sinkhorn algorithm to perform the forward inference in \textit{soft KNN}. The Sinkhorn algorithm is useful in entropy-regularized optimal transport problems thanks to its computational effort reduction. The \textit{soft KNN} has $\mathcal{O}(n)$ complexity, making it scalable and allows us to safely apply it to more computationally expensive problems. 

The values of \textit{soft KNN} hyperparameters $\sigma$ and $L$ are also presented in Table \ref{tab:soft-knn}. We utilize the continuous character of output vector to weight the ensemble predictions. It is worth noting that we additionally set the threshold of the minimum allowed \textit{soft KNN} score to 0.3. It means every element in $\boldsymbol\gamma$ lower than 0.3 is reduced to 0. We reject such elements because they are mostly the result of non-converged optimization and do not carry important information. In this way, we additionally secure the optimization result to be as representative as possible.

\subsection{Complexity and ablations.}\label{appendix:section:complexity}

\begin{figure}[!ht]
\centering
\begin{minipage}{.45\textwidth}
  \centering
  \includegraphics[width=\linewidth]{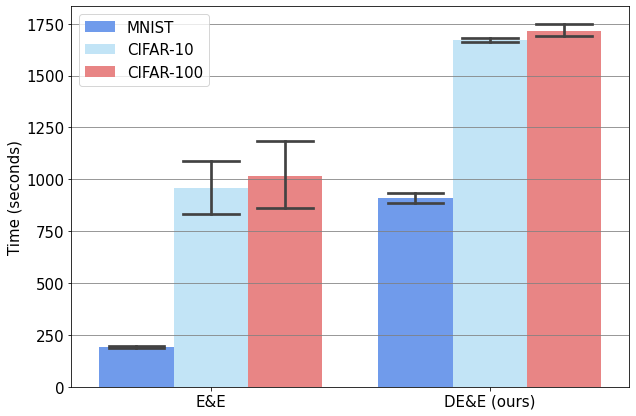}
  \captionof{figure}{Training time for DE\&E ensemble model (128 classifiers). Results for 10-split MNIST, 10-split CIFAR-10 and 20-split CIFAR-100 are shown.}
  \label{fig:time}
\end{minipage}
\hspace{1cm}
\begin{minipage}{.46\textwidth}  \centering
  \includegraphics[width=\linewidth]{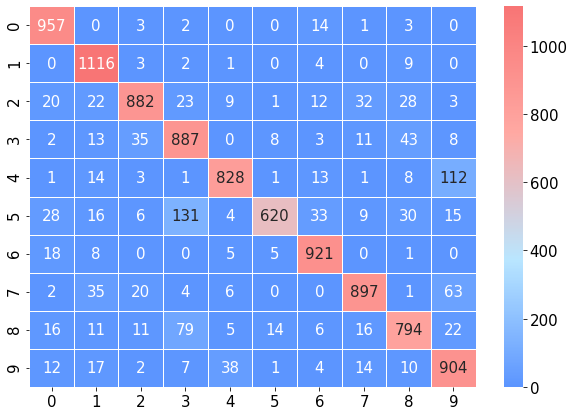}
  \captionof{figure}{Confusion matrix for DE\&E ensemble model (128 classifiers) evaluated on the 10-split MNIST.}
  \label{fig:cm}
\end{minipage}%
\end{figure}

\begin{figure}[!ht]
\centering
\begin{minipage}{.46\textwidth}  \centering
  \includegraphics[width=\linewidth]{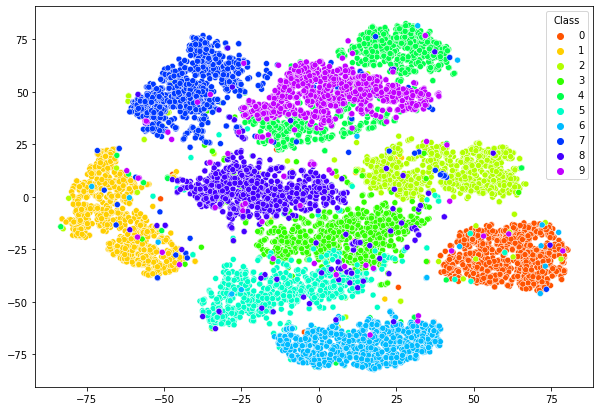}
  \captionof{figure}{TSNE visualization of MNIST data set encodings (VAE encoder). 10000 examples are shown.}
  \label{fig:tsne-mnist}
\end{minipage}%
\hspace{1cm}
\begin{minipage}{.46\textwidth}
  \centering
  \includegraphics[width=\linewidth]{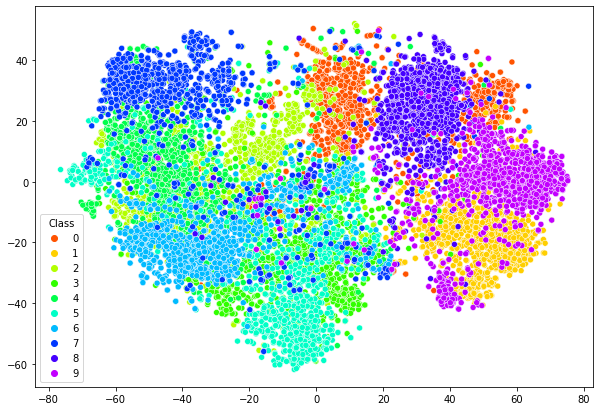}
  \captionof{figure}{TSNE visualization of CIFAR-10 data set encodings (BYOL ResNet-50). 10000 examples are shown.}
  \label{fig:tnse-cifar10}
\end{minipage}
\end{figure}

The machine we used had 128 GB RAM, an Intel Core i9-11900 CPU, and an NVIDIA GeForce RTX 3060 GPU with 12GB VRAM. Every experiment was performed using the GPU. The comparison in training time between E\&E and DE\&E models is shown in Figure \ref{fig:time}. For all evaluated data sets, the training time of our model was higher than the time to train the reference method. The results vary between data sets. In case of MNIST, the time to train fully differentiable neural architecture was about four times longer than the E\&E. The difference is also noticeable in much more difficult cases (CIFAR-10 and CIFAR-100).

We observed several important aspects that affect the performance of the whole model. Firstly, since the weak learners are single-layer neural networks, the entire feature extraction process relies on a pre-trained encoder that strongly influences the upper bounds of classification accuracy. Models in the ensemble only learn the feature mapping to the class, reducing the complexity and thus reducing the forgetting phenomenon.
% For this reason, the feature extractor plays a key role that strongly influences the possibility of achieving satisfactory performance. Thus, the acquisition of good quality encodings provides the ensemble with the conditions to achieve high classification performance. In contrast, a poorly prepared encoder degrades the discriminative capabilities of the ensemble at the start.
The conducted research indicates that with the same features, the proposed method is able to obtain higher results than the reference methods. Thus, the described problem can be partially overcome by using the architecture proposed in this paper, since the decrease in feature quality has the potential to be compensated by skillful use of other mechanisms.

% This same phenomenon accounts for the model's most common errors. We observed that the most frequently confused classes are those most visually similar to each other.
%Figure [X] shows the confusion matrix for model [describe] prediction on the 10-split MNIST test set. 
% Across all evaluations on the MNIST data set the most errors were made between classes 3 and 5 and also 4 and 9. In contrast, classes 6 and 7 were misclassified only a few times. This indicates a limitation posed by the feature extractor, whose output is not detailed enough to sufficiently distinguish between highly similar examples. 
%In addition, we also applied the T-SNE nonlinear dimension reduction algorithm, whose result is shown in Figure [X].
% The obtained results suggest a clear difficulty in distinguishing the mentioned classes from each other. This phenomenon is compounded on the CIFAR-10 and CIFAR-100 collections, where the detail of examples is higher and they require much more accurate features. Thus, the use of a more accurate encoder would probably result in an increased classification accuracy performance.

An apparent drawback of the proposed neural model is the increased training time.% as seen in Figure [X].
The introduction of a differentiable \textit{soft KNN} layer resulted in additional computational effort that clearly impacted the time complexity of the model. \textit{soft KNN} is based on an approximation of the Entropic Optimal Transport problem solution, which is typically computationally expensive. The convergence rate of the algorithm depends on its hyperparameters that should be tuned so that the assumed error tolerance is reached as quickly as possible. Apart from optimization problems, aggregating weak learner predictions is computationally expensive, too. Unlike the reference method, where classifiers from the nearest neighborhood are trained in complete isolation, our method computes the output of all weak learners during each forward pass. However, only the learners that belong to the nearest neighbor group are selected for update (the output of the \textit{soft KNN} layer is used for the selected classifiers to be updated). The adopted weighting procedure makes it possible to simultaneously eliminate the predictions of classifiers that are not nearest neighbors, as well as give appropriate proportions to the predictions that qualify for the group of nearest neighbors. In this case the weighting is not done for $\kappa$ neighbors but for all $N$ classifiers in the ensemble, which is very time-consuming for large $N$. Here, we see the field for improvement and more efficient use of the differentiable model capabilities for future work.

We observed that the cause of DE\&E prediction errors is not much different than errors made by other models. The most frequently confused classes are those most visually similar to each other.
Figure \ref{fig:cm} shows the confusion matrix for DE\&E (128 classifiers) prediction on the 10-split MNIST test set. Across all evaluations on the MNIST data set, the highest number of errors were made between classes 3 and 5, and also 4 and 9. In contrast, classes 6 and 7 were misclassified only a few times. This indicates a limitation posed by the feature extractor, whose output is not detailed enough to sufficiently distinguish between highly similar examples. 
In addition, we also applied the T-SNE nonlinear dimensionality reduction algorithm into MNIST and CIFAR-10 encodings. The output is shown in Figures \ref{fig:tsne-mnist} and \ref{fig:tnse-cifar10}.
The obtained results suggest a clear difficulty in distinguishing the mentioned classes from each other. This phenomenon is compounded on the CIFAR-10 and CIFAR-100 data sets, where the detail of examples is higher and they require much more accurate features. Thus, the use of a more accurate encoder would probably result in an increased classification accuracy performance.

\subsection{Other experiments.}

\subsubsection{Forgetting}

\begin{figure}[!tp]
    \centering
    \includegraphics[width=\textwidth]{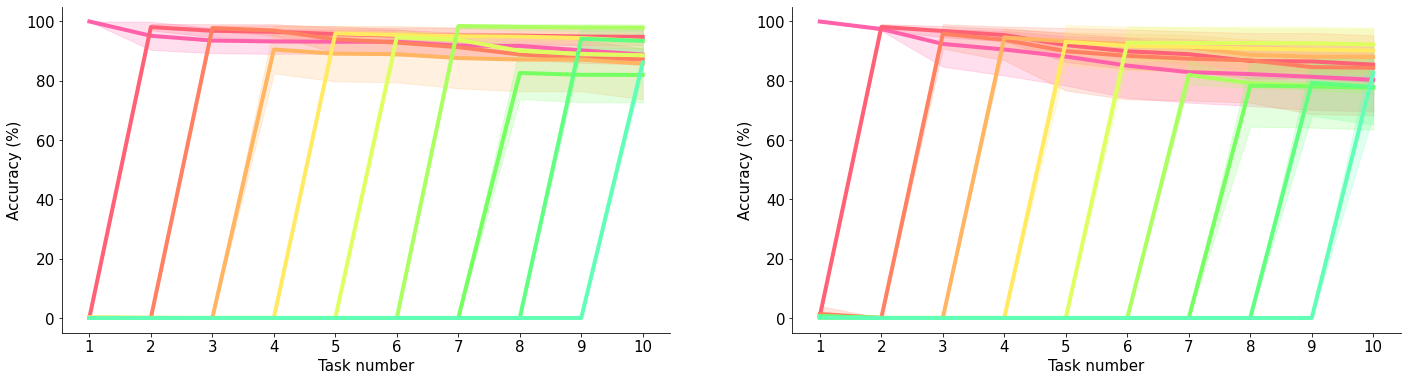}
    \caption{Accuracy (\%) performance across training tasks by DE\&E ensemble of 1024 (left) and 16 (right) classifiers on 10-split MNIST. The smaller ensemble accuracy decreases much more for consecutive tasks due to no possibility of classifiers specialization.}
    \label{fig:training}
\end{figure}

In Tables \ref{tab:forgetting-5split-mnist}, \ref{tab:forgetting-10split-mnist}, \ref{tab:forgetting-5split-cifar10}, and \ref{tab:forgetting-10split-cifar10} we also show the forgetting rates for ensembles trained on MNIST and CIFAR-10 data sets. The forgetting measure fluctuates greatly across evaluations, especially on smaller ensembles. Additionally, the forgetting rate is influenced by class order. We observed that lower forgetting rates are particularly noticeable in the simpler MNIST cases. Importantly, forgetting is always reduced in large ensembles (1024 classifiers) using our method. As can be seen in Figure \ref{fig:training}, the larger ensemble learns each task better
than smaller one. It also has a stable performance on the already seen tasks. The smaller ensemble acquires less knowledge on subsequent tasks and forgets much more.

\begin{table}[!ht]
    \caption{Forgetting (\%) and standard deviation (5 runs) for various ensemble sizes evaluated on 5-split MNIST.}
    \label{tab:forgetting-5split-mnist}
    % \fontsize{6}{6}\selectfont
    % \renewcommand{\arraystretch}{1.6}
    \begin{tabular*}{\textwidth}{@{\extracolsep{\fill}}lccc@{}}
    \toprule
                        & 16     & 128    & 1024  \\ \midrule
    E\&E   & $9.07 \pm 2.85$           & $\mathbf{5.18 \pm 1.98}$          & $5.61 \pm 0.66$          \\
    DE\&E (ours)  & $\mathbf{6.96 \pm 1.95}$           & $5.20 \pm 1.51$          & $\mathbf{4.90 \pm 1.54}$      \\ \bottomrule
    \end{tabular*}
\end{table}

\begin{table}[!ht]
    \caption{Forgetting (\%) and standard deviation (5 runs) for various ensemble sizes evaluated on 10-split MNIST.}
    \label{tab:forgetting-10split-mnist}
    % \fontsize{6}{6}\selectfont
    % \renewcommand{\arraystretch}{1.6}
    \begin{tabular*}{\textwidth}{@{\extracolsep{\fill}}lccc@{}}
    \toprule
                        & 16     & 128    & 1024  \\ \midrule
    E\&E   & $8.13 \pm 2.57$           & $8.16 \pm 1.90$          & $5.26 \pm 0.52$ \\
    DE\&E (ours)    & $\mathbf{6.71 \pm 2.18}$           & $\mathbf{6.04 \pm 1.81}$          & $\mathbf{1.50 \pm 2.32}$  \\ \bottomrule
    \end{tabular*}
\end{table}

\begin{table}[!ht]
    \caption{Forgetting (\%) and standard deviation (5 runs) for various ensemble sizes evaluated on 5-split CIFAR-10.}
    \label{tab:forgetting-5split-cifar10}
    % \fontsize{6}{6}\selectfont
    % \renewcommand{\arraystretch}{1.6}
    \begin{tabular*}{\textwidth}{@{\extracolsep{\fill}}lccc@{}}
    \toprule
                        & 16     & 128    & 1024  \\ \midrule
    E\&E   & $\mathbf{17.28 \pm 3.79}$           & $15.38 \pm 4.38$          & $11.88 \pm 1.40$          \\
    DE\&E (ours)  & $18.40 \pm 3.73$           & $\mathbf{12.63 \pm 1.89}$          & $\mathbf{11.78 \pm 3.51}$      \\ \bottomrule
    \end{tabular*}
\end{table}

\begin{table}[!ht]
    \caption{Forgetting (\%) and standard deviation (5 runs) for various ensemble sizes evaluated on 10-split CIFAR-10. Result for the E\&E with 128 classifiers is not reported due to technical reasons during evaluation.}
    \label{tab:forgetting-10split-cifar10}
    % \fontsize{6}{6}\selectfont
    % \renewcommand{\arraystretch}{1.6}
    \begin{tabular*}{\textwidth}{@{\extracolsep{\fill}}lccc@{}}
    \toprule
                        & 16     & 128    & 1024  \\ \midrule
    E\&E   & $\mathbf{9.89 \pm 11.49}$           & -          & $13.84 \pm 2.64$ \\
    DE\&E (ours)    & $21.58 \pm 4.32$           & $\mathbf{15.77 \pm 3.17}$          & $\mathbf{13.82 \pm 2.44}$  \\ \bottomrule
    \end{tabular*}
\end{table}

\subsubsection{Trainable keys}\label{appendix:section:trainable-keys}

\begin{table}[!ht]
    \caption{Impact of trainable keys on model accuracy (\%). Experiments are performed on the MNIST data set. Means and standard deviations for 5 runs are shown.}
    \label{tab:trainable-keys}
    % \fontsize{8}{8}\selectfont
    % \renewcommand{\arraystretch}{1.6}
    \centering
        \begin{tabular*}{\textwidth}{@{\extracolsep{\fill}} lcccc @{}}
    \toprule
                        & \multicolumn{2}{c}{5-split} & \multicolumn{2}{c}{10-split} \\  \cmidrule(l){2-5} 
                        & $N = 16$          & $N = 128$         & $N = 16$            & $N = 128$           \\ \midrule
    DE\&E    & $85.20 \pm 0.46$  & $\mathbf{87.80 \pm 0.47}$  & $84.19 \pm 1.00$    & $\mathbf{87.54 \pm 0.24}$   \\
    DE\&E + trainable keys & $\mathbf{85.30 \pm 0.51}$  & $87.47 \pm 0.22$  & $\mathbf{84.64 \pm 0.69}$    & $87.23 \pm 0.12$  \\ \bottomrule
    \end{tabular*}
\end{table}

Table \ref{tab:trainable-keys} shows the effect of trainable keys on ensemble accuracy. For keys optimization, we used the Adam optimizer and a learning rate of 0.0005. As for previous experiments, a positive effect on accuracy is seen for a small ensemble of 16 classifiers. We observe this in both the 5-split and 10-split setups on the MNIST data set. In contrast, for an ensemble of 128 classifiers, key optimization leads to a decrease in prediction quality. We conclude that with a fewer number of keys it is much more difficult to allow them to specialize. Giving the keys a degree of freedom may lead to an adjustment of their arrangement. However, when the number of keys is higher and the space is covered more densely, correcting them is ineffective.

\subsubsection{Large ensembles and experiments on CIFAR-100}\label{appendix:section:large-and-100}

As shown in Table \ref{tab:main-results-cifar100}, in case of the CIFAR-100 and the largest ensemble, no significant improvement was noted for 20-split, but for 100-split the improvement was about 9\%. But for smaller ensembles we do not observe significant gains over the E\&E (Table \ref{tab:cifar-ensemble}).
This clearly shows that ensembles too small in size (according to problem difficulty) does not give a chance for improvement with the proposed techniques. When there are too few models, the collective intelligence is suppressed by voting noise. However, it is important to note that increasing ensemble size does not always result in improved performance. As we have shown before, even an ensemble of 16 classifiers can perform much more accurate on the MNIST data set just by making proper use of the specialization mechanism. Small ensembles benefit most from the proposed voting method, because their each vote has a relatively greater impact on the decision than for larger models. Thus, a streamlined voting procedure results in proportionally more meaningful effects.

In case of the largest ensemble (1024 classifiers) we observed slight improvement over the E\&E method in all setups evaluated (Table \ref{tab:main-results}). Our method achieves higher accuracy in both 5-split and 10-split scenarios for MNIST and CIFAR-10 datasets.

\begin{table}[!ht]
    \caption{Accuracy (\%) and standard deviation for models evaluated on CIFAR-100 data set. Both E\&E and DE\&E ensembles consist of 1024 classifiers.}
    \label{tab:main-results-cifar100}
    % \fontsize{6}{6}\selectfont
    % \renewcommand{\arraystretch}{1.6}
    \centering
    \begin{tabular*}{0.53\textwidth}{lcc}
    \toprule
                        
                        & 20-split      & 100-split     \\ \midrule
    E\&E   & $\mathbf{40.34 \pm 0.58}$             & $31.60 \pm 8.12$             \\
    DE\&E (ours) & $39.72 \pm 8.25$            & $\mathbf{40.57 \pm 6.00}$             \\ \bottomrule
    \end{tabular*}
\end{table}

\begin{table}[!ht]
    \caption{Accuracy (\%) and standard deviation (5 runs) for smaller ensembles on CIFAR-100.}
    \label{tab:cifar-ensemble}
    % \fontsize{6}{6}\selectfont
    % \renewcommand{\arraystretch}{1.5}
    \centering
        \begin{tabular*}{0.5\textwidth}{lcc}
    \toprule
                        % \\ \cmidrule(l){2-3} 
                        & $N = 16$          & $N = 128$          \\ \cmidrule(l){1-3} 
    E\&E   & $7.04 \pm 1.56$  & $30.52 \pm 0.74$  \\
    DE\&E (ours) & $\mathbf{7.19 \pm 0.54}$  & $\mathbf{30.58 \pm 1.06}$ \\ \bottomrule
    \end{tabular*}
\end{table}

\begin{table}[!ht]
    \caption{Accuracy (\%) and standard deviation (5 runs) for models evaluated on MNIST and CIFAR-10 data sets. Both E\&E and End-to-End ensembles consist of 1024 classifiers.}
    \label{tab:main-results}
    % \fontsize{6}{6}\selectfont
    % \renewcommand{\arraystretch}{1.6}
    \begin{tabular*}{\textwidth}{@{\extracolsep{\fill}}lcccc@{}}
    \toprule
                        & \multicolumn{2}{c}{MNIST} & \multicolumn{2}{c}{CIFAR-10} \\ \cmidrule(l){2-5} 
                        & 5-split     & 10-split    & 5-split      & 10-split      \\ \midrule
    E\&E   & $89.13 \pm 0.23$           & $89.53 \pm 0.31$          & $66.27 \pm 0.77$            & $67.41 \pm 0.80$            \\
    DE\&E (ours) & $\mathbf{89.45 \pm 0.18}$           & $\mathbf{90.00 \pm 0.21}$          & $\mathbf{67.44 \pm 0.67}$            & $\mathbf{67.57 \pm 1.06}$           \\ \bottomrule
    \end{tabular*}
\end{table}

\subsection{Related work}

\paragraph{Continual learning methods.} Currently, methods with a memory buffer such as GEM \cite{GEM}, A-GEM \cite{AGEM} or MIR \cite{MIR} usually achieve the highest performance on benchmark tasks using traditional data sets \cite{RecencyBias}. Because past samples are stored in memory and repeated multiple times during training, forgetting is reduced by constantly refreshing the knowledge acquired in the past. It has been shown that even a very small number of stored examples can significantly reduce network performance degradation \cite{TinyEpisodicMemory}. In addition to methods with a memory buffer, a very wide group of approaches based on parameter regularization exists. The most popular ones include EWC \cite{EWC} or LWF \cite{LWF}. When receiving a new dose of knowledge, these methods attempt to influence the model parameter updating procedure to be minimally invasive. The lack of a memory buffer results in a smaller memory overhead, but the price paid for this means lower efficiency. Another group of methods are approaches based on the expansion of network parameters like PackNet \cite{mallya2018packnet} or Progressive Neural Networks \cite{ProgressiveNeuralNetworks}. However, the main limitation of those methods is the significant increase in the number of parameters during training.

\paragraph{Ensemble methods.} Ensemble methods are widespread in the field of deep learning \cite{yang2021survey, cao2020ensemble, li2022novel}. Ensemble techniques have also been used successfully in the field of continual learning, as evidenced by the presence of methods such as BatchEnsemble \cite{wen2020batchensemble} or CN-DPM \cite{CN-DPM}. Other contributions present in literature \cite{doan2022efficient} tend to focus strongly on improving model performance rather than increasing model efficiency. Furthermore, ensemble approaches can also be used indirectly through dropout \cite{Dropout} or weights aggregation \cite{wortsman2022model}.

\paragraph{KNN.} The K Nearest Neighbors algorithm is currently the most frequently used among a variety of machine learning techniques \cite{KNNPopular}. Despite its very high computational cost for a large number of examples, it is still often used as a baseline in various classification and regression problems. However, KNN is characterized by a lack of differentiability, effectively blocking the exploitation of its advantages in gradient training based neural architectures. This problem has attracted the attention of researchers in recent years, leading to the development of several methods that allow for approximating the output of a conventional KNN guaranteeing the ability to calculate the needed derivatives. One of the first such approaches was the use of Differentiable Boundary Trees \cite{Diff-knn-bounded-trees}, where the authors proposed a custom cost function associated with a tree's prediction. Another important work is the introduction of continuous deterministic relaxation of KNN \cite{N3Block}, which can be used directly as a neural network layer ($N^3$ block). This method has shown effectiveness in both classification and image denoising problems. The most recent approach \cite{Diff-KNN-SOFT} (used by us) that solves the Entropic Optimal Transport problem. Importantly, the introduced form of optimization also allowed to improve the classification results, this time on the CIFAR-10 data set. However, as for each method mentioned above, the computational cost behind accuracy gains is quite significant.

% \paragraph{E\&E limitations}

%The main problem of E\&E approach is the presence of a non-differentiable $k$ nearest neighbors (KNN) module, whose task is to select $k$ specialized classifiers that can handle the input. This separates the ensemble from the rest of the architecture, paving the way for fully differentiable end-to-end learning. This limitation greatly restricts its further development and practical usage, because advances in the field of neural networks and deep learning clearly show the benefits of training very complex models as a whole using gradient algorithms \cite{kotary2021end}.

\end{document}